\documentclass{Interspeech2024}

\usepackage{graphicx} 
\usepackage[utf8]{inputenc} 
\usepackage{hyperref}
\usepackage[finalnew]{trackchanges}
\usepackage{textalpha} 

\usepackage{multirow}

\interspeechcameraready

\title{The Greek podcast corpus: Competitive speech models for low-resourced languages with weakly supervised data}

\name[affiliation={1}]{Georgios}{Paraskevopoulos}
\name[affiliation={1}]{Chara}{Tsoukala}
\name[affiliation={1}]{Athanasios}{Katsamanis}
\name[affiliation={1}]{Vassilis}{Katsouros}

\address{
  $^1$Institute for Speech and Language Processing, Athena Research Center, Athens, Greece}
\email{\{g.paraskevopoulos,chara.tsoukala,nkatsam,vsk\}@athenarc.gr}


\begin{document}

\maketitle

\begin{abstract}
The development of speech technologies for languages with limited digital representation poses significant challenges, primarily due to the scarcity of available data. This issue is exacerbated in the era of large, data-intensive models. Recent research has underscored the potential of leveraging weak supervision to augment the pool of available data. In this study, we compile an 800-hour corpus of Modern Greek from podcasts and employ Whisper large-v3 to generate silver transcriptions. This corpus is utilized to fine-tune our models, aiming to assess the efficacy of this approach in enhancing ASR performance.
Our analysis spans 16 distinct podcast domains, alongside evaluations on established datasets for Modern Greek. The findings indicate consistent WER improvements, correlating with increases in both data volume and model size. Our study confirms that assembling large, weakly supervised corpora serves as a cost-effective strategy for advancing speech technologies in under-resourced languages. 
\end{abstract}

\section{Introduction}

Speech technologies, specifically Automatic Speech Recognition (ASR), have evolved in recent years transitioning towards end-to-end systems~\cite{pmlr-v32-graves14,conneau2021unsupervised}. The good performance achieved by neural ASR systems has accelerated research and given rise to innovative industry applications. 

Building upon these advancements, the field has seen significant progress in developing more efficient and effective end-to-end systems. 
However, this evolution has been mostly focused on languages with abundant digital resources, primarily English, leading to a stark imbalance in technology accessibility and performance across languages. The introduction of massively multilingual speech recognition models~\cite{conneau2021unsupervised,radford2023robust} has begun to address this disparity, offering a pathway to include digitally underrepresented languages. Despite these efforts, there remains an urgent need for creating and expanding new corpora specifically tailored to these languages, but current efforts are costly and fragmented.

In addressing the challenge of developing comprehensive and dynamically updated corpora for the data-intensive demands of contemporary neural models, self-training~\cite{kahn2020self} emerges as a cost-effective strategy. This method has gained more relevance with current advancements in model performance, improving its effectiveness for augmenting language data resources. Notably, self-training has been effectively utilized for low-resource languages, as demonstrated by Ragni et al.~\cite{ragni2014data} within a hybrid DNN-HMM framework and by Singh et al.~\cite{singh23b_interspeech} through an iterative pseudo-labeling scheme. The development of Distil-Whisper has further validated the potential and efficiency of self-training techniques~\cite{gandhi2023distil}, where efficient versions of Whisper have been crafted using pseudo-labeled corpora with minimal impact on model performance.

To address the scarcity of data, podcasts can be used to create diverse and extensive speech corpora. In 2020, Clifton et al. \cite{clifton2020spotify} released the Spotify Dataset, which consists of 100,000 episodes and nearly 60,000 hours of English speech. This dataset was later augmented to include Portuguese podcasts \cite{garmash2023cem}. Despite its initial value, the Spotify Dataset has ceased maintenance and is no longer accessible. Among other significant contributions to the field, the People's speech dataset \cite{peoplespeechdataset2021} and Gigaspeech \cite{GigaSpeech2021} have been developed using podcasts, yet both predominantly focus on English. To the best of our knowledge, there are no podcast corpora tailored to low-resource languages. The potential benefits of establishing such resources are substantial, as, combined with self-training and pseudo-labeling, such corpora can be used to train larger speech models.

In this work, we assemble a podcast corpus for Modern Greek (Greek Podcast Corpus; GPC), spanning $16$ distinct domains to facilitate multi-domain evaluation. The  WhisperX pipeline~\cite{bain23_interspeech} is used for corpus segmentation and transcription, in conjunction with Whisper large-v3 \cite{radford2023robust}, a cutting-edge, massively multilingual ASR model. We collect an untranscribed pre-training corpus with $3124$ hours of speech, and a transcribed ASR corpus divided into training, validation, and test sets, with $623$, $4$, and $13$ hours of speech respectively. The training, validation, and test splits are evenly stratified across the $16$ domains. Additionally, we identify two single-speaker podcasts with high-quality audio, suitable for TTS training purposes. To gauge the efficacy of self-training in enhancing ASR capabilities, we fine-tune Whisper-small and Whisper-medium on the newly constructed ASR corpus using varying quantities of speech. Our evaluation includes standard corpora, not included in the fine-tuning set. Furthermore, we conduct a thorough analysis across different domains using the podcast test set. The results demonstrate the positive impact of weakly-supervised fine-tuning on Greek ASR performance, with observed improvements scaling with model size and training data quantity.

Our contributions are a) the creation of a large, multi-domain podcast corpus for Modern Greek, which can be easily expanded to thousands of speech hours, and b) extensive multi-domain and out-of-training evaluation demonstrating the effectiveness of utilizing weakly supervised corpora. 
The trained model checkpoints, as well as recipes to recreate the corpora and models, 
are publicly available\footnote{\url{https://github.com/georgepar/greek_podcasts_asr}} for research purposes, aligning with the AI Act (Art. 2, Par. 6)\footnote{\url{https://www.europarl.europa.eu/doceo/document/TA-9-2024-0138_EN.pdf}}. 
Note, we do not redistribute the original data, rather we provide scripts for corpus reproduction.

\section{Background: The WhisperX pipeline}

The collected podcasts are available in long-form audio format, but current ASR pipelines require segmented and aligned audio for training. In the case of Whisper, the model can handle samples with maximum duration $T_0=30$ seconds. 
To segment and transcribe the collected speech corpus we utilize the WhisperX pipeline~\cite{bain23_interspeech}. This pipeline yields a list of speech segments of roughly $30$-second duration, paired with the obtained transcriptions. The pipeline includes four steps:

\begin{table}
\caption{Analysis of the collected podcasts in the GPC pre-training corpus per category.}
\label{tab:podcast_domain_descr}
\centering
\begin{tabular}{lrrr}
\hline
\textbf{Domain} & \textbf{Total hours} & \textbf{\#podcasts} & \textbf{\#episodes} \\
\hline
Arts & $368$ &  $24$ & $1977$ \\
Business & $151$ & $12$ & $539$\\
Comedy & $249$ & $23$ & $667$\\
Education & $262$ & $35$ & $1003$ \\
HealthFitness & $165$ & $20$ & $899$ \\
History & $126$ & $12$ & $666$ \\
KidsFamily & $60$ & $8$ & $214$\\
Leisure & $243$ & $10$ & $669$\\
Music & $217$ & $10$ & $2974$ \\
News & $176$ & $10$ & $2374$ \\
Science & $114$ & $8$ & $147$\\
SocietyCulture & $275$ & $34$ & $1159$ \\
Sports & $258$ & $23$ & $873$ \\
Technology & $151$ & $8$ & $1037$ \\
TrueCrime & $122$ & $7$ & $230$ \\
TVFilm & $187$ & $11$ & $409$ \\
\hline
Total & $3124$ & $255$ & $15837$\\
\hline
\end{tabular}
\end{table}

\begin{enumerate}
\item \textbf{Voice Activity Detection (VAD):} The input audio is split 
into segments that contain speech (active), and inactive segments that contain silence, music, noise, etc. The pyannote VAD model~\cite{plaquet23_interspeech,bredin23_interspeech} is chosen for this step.
\item \textbf{Cut and Merge:} Following VAD, segments containing active speech are further processed to fit the input constraints of the Whisper model. Longer segments are divided into smaller parts at points of low VAD scores. Additionally, adjacent segments of very short duration are combined until they reach a maximum length of $30$ seconds.
\item \textbf{Transcription:} The core of the pipeline leverages the whisper-large-v3 model\footnote{\url{https://huggingface.co/openai/whisper-large-v3}}, trained on a vast dataset comprising $1$ million hours of weakly labeled audio and an additional $4$ million hours of pseudo-labeled audio. This model, consisting of $1550$M parameters, yields a $10-20\%$ improvement over the previous version.
\item \textbf{Alignment:} The final step involves obtaining word-level timestamps, using a phoneme-level model for forced alignment. We use a version of XSLR-53\footnote{\url{https://huggingface.co/jonatasgrosman/wav2vec2-large-xlsr-53-greek}} \cite{grosman2021xlsr53-large-greek} which has been fine-tuned on the Greek parts of Common Voice (CV)~\cite{commonvoice:2020} and CSS10~\cite{park19c_interspeech}. Bain et al.~\cite{bain23_interspeech} have demonstrated this to yield more accurate and robust timestamps than Whisper.
\end{enumerate}

\section{The Greek Podcast Corpus}

\subsection{Data collection and preprocessing}

The data collection consists of two steps and is built using available open-source tools. First, we build a web crawler to collect RSS feeds of relevant podcasts~\footnote{We collect RSS feeds from \url{https://podcastaddict.com/?lang=el}. This pipeline can be run for any of the $27$ languages included in this website.}. RSS (Really Simple Syndication) is a web feed format used to publish frequently updated content, such as blog entries, news headlines, or podcasts, in a standardized, machine-readable format.

After collecting the RSS feeds, the audio can be downloaded using an open-source ``podcatching'' tool.
After the audio is downloaded, we convert it to WAV format, single-channel, at $16$ kHz sampling rate. 
At the time of writing, this process has yielded $3124$ hours of audio, which can be readily used as a pre-training corpus. Table~\ref{tab:podcast_domain_descr} breaks down the collected corpus statistics per category.

Additionally, during data collection we manually identified $3$ podcasts by $2$ speakers ($24$ and $123$ hours respectively), with clean, single-speaker speech, that may be useful for training TTS models. We include them in the GPC corpus for future use by the research community.

\subsection{Subcorpora and splits}

\begin{figure}
    \centering
    \includegraphics[width=.25\textwidth]{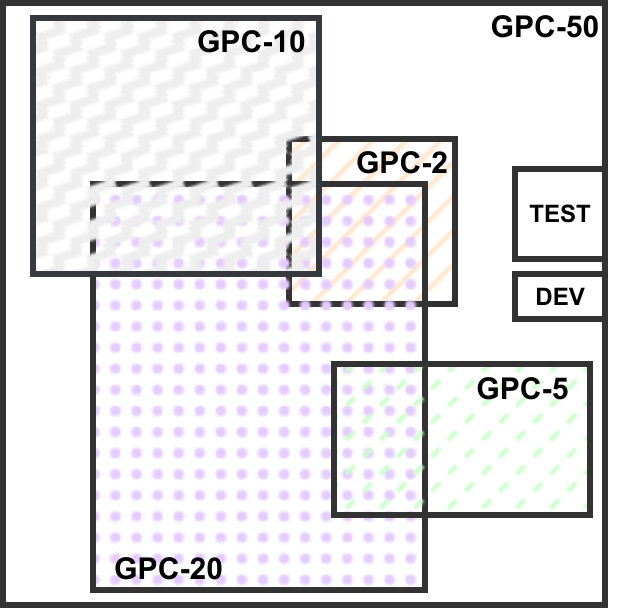}
    \caption{Visualization of the sampling procedure for the subsets of the GPC-50 corpus.}
    \label{f:gpc-subsets}
\end{figure}

To construct the ASR corpus, we categorize the podcasts and exclude those categories that either have only a minimal number of podcasts or predominantly feature speech in ``purist" Greek, which is an archaic variant of Modern Greek. This process results in a selection of 16 diverse categories. Following this, we randomly select $50$ hours from each category to assemble a stratified and varied ASR corpus totaling $800$ hours of audio, denoted as GPC-50 for the rest of this paper. Within this corpus, we designate a test split and a validation split, allocating $1$ hour per category (for a total of 16 hours) and $15$ minutes per category (for a total of $4$ hours), respectively. The rest of the audio is allocated to the GPC-50 training set.
We also create smaller training subsets from the GPC-50 training set by sampling $20$, $10$, $5$, and $2$ hours from each category, resulting in the GPC-20, GPC-10, GPC-5, and GPC-2 subsets. These contain total audio durations of $320$, $160$, $80$, and $32$ hours, respectively. Fig.~\ref{f:gpc-subsets} illustrates the created subsets.

\subsection{Automatic Transcription}

GPC-50 is transcribed and segmented utilizing the WhisperX pipeline. Following this, we implement two post-processing procedures. First, we eliminate segments where Whisper produces hallucinations, commonly occurring during music sections not filtered out by VAD. These can be readily identified through the ``Υπότιτλοι AUTHORWAVE" (trans. ``AUTHORWAVE subtitles") caption. 
Subsequently, due to the phoneme-level aligner's 
monolingual design, segments containing 
code-switched English-Greek phrases 
may present inaccurate timestamps and were consequently removed. This issue 
mainly arises when English text appears at the beginning or end of a transcript, for instance, 
``ακολουθήστε με στο Instagram" (trans. ``follow me on Instagram"). 
We observe an approximate $20\%$ reduction in data size due to the inactive segment filtering by VAD and our post-processing. This leaves $623$, $3.8$, and $13$ hours of useable training, validation, and test data respectively.




\section{Experimental setup}

\subsection{Experimental details}

Given our limited computing budget, we fine-tune whisper-small and whisper-medium (with $244$M and $769$M parameters respectively) on GPC for speech recognition for $4$ epochs. We perform fine-tuning for different amounts of training data on GPC-\{50,20,10,5,2\} training splits. For training, we use a linear warmup schedule of $500$ steps, the batch size is set to $16$ samples per device and accumulate gradients every $4$ training steps. The learning rate is set to $10^{-5}$ and we set the maximum generation length to $225$. We use $16$-bit floating point arithmetic and the standard ``sdpa'' attention implementation during training.
Fine-tuning runs on $1$ RTX 3090 GPU with $24$ GB VRAM for all scenarios except for fine-tuning whisper-medium on GPC-20 and GPC-50, where we use $2$ RTX 3090 GPUs and halve the gradient accumulation steps to maintain an equivalent effective batch size across experiments.
During decoding, we normalize the reference and transcriptions and report the Word Error Rate (WER) over the samples of the respective test sets.

\subsection{Datasets}

To assess the impact of continued training on weakly supervised data, we evaluate the fine-tuned models using the standard test sets of four diverse speech corpora in Modern Greek. These corpora are used exclusively for evaluation purposes. Although we do not utilize these corpora for fine-tuning, we cannot confirm whether they have been encountered during Whisper's pre-training phase. The four benchmark corpora are the following:

\noindent\textbf{Fleurs}~\cite{conneau2023fleurs} is a multilingual speech corpus targeted for few-shot evaluation of speech models. It contains $3000$ sentences by English Wikipedia, read by male and female participants, across $104$ languages, resulting in $1400$ hours of parallel speech. The Greek subset of Fleurs contains 
a total of $13$ hours of speech, with the test set containing $2$ hours of speech.

\noindent\textbf{Common Voice (CV)}~\cite{commonvoice:2020} version 11.0, accessed on April 27, 2022 is a crowdsourced, multilingual corpus of dictated speech, developed by Mozilla through a web or iPhone app. Contributors read prompts sourced from public domains, e.g., books and Wikipedia, limited to 15 words. 
The dataset, comprising $18$ hours of validated speech from $325$ contributors aged $19$ to $59$ with male and female voices, is split into the standard train, development, and test sections. The CV test set contains $2$ hours of speech. 

\noindent\textbf{Logotypografia (LG)}~\cite{digalakis03_eurospeech} is a pioneering corpus for Large Vocabulary Continuous Speech Recognition (LVCSR) in Greek, featuring $33136$ newscast utterances ($72$ hours of speech) from $125$ speakers ($55$ males, $70$ females) associated with the well-known ``Eleftherotypia'' newspaper in Greece. These recordings, captured under diverse acoustic settings, include sessions in a soundproof room, a quiet room, and an office environment. On average, utterances last $7.8$ seconds. The dataset's transcriptions are rich, including speech and non-speech sounds (e.g., \<cough\>), Greek words in lowercase with stress marks, and numbers spelled out. We remove 
non-speech events from the transcriptions during evaluation.
The LG test set contains $9$ hours.

\begin{table}
\caption{Evaluation on the GPC-50 test set.}
\label{tab:podcast-50h}
\centering
\begin{tabular}{ccccr}
\toprule
\textbf{Dataset} & \textbf{Model} & & \textbf{Finetuning corpus} & \textbf{WER} \\
\toprule
\multirow{5}{*}{GPC-50} & small & & - & $42.87$ \\
       &  small & & GPC-50 & $\underline{12.16}$ \\
       & medium & & - & $24.22$ \\
       & medium & & GPC-50 & $\mathbf{9.95}$ \\
       & large-v2 & & - & $17.27$ \\
\bottomrule
\end{tabular}
\end{table}

\begin{table}
\caption{Evaluation on standard out-of-training Greek corpora.}
\label{tab:50h_dataset_model_overview}
\centering
\begin{tabular}{ccccr}
\toprule
\textbf{Dataset} & \textbf{Model} & & \textbf{Finetuning corpus} & \textbf{WER} \\
\toprule
\multirow{6}{*}{fleurs} & small & & - & $33.99$ \\
       &  small & & GPC-50 & $\underline{16.78}$ \\
       & medium & & - & $19.45$ \\
       & medium & & GPC-50 & $\underline{14.03}$ \\
       & large-v2 & & - & $12.95$ \\
       & large-v3 & & - & $\mathbf{11.02}$ \\
\midrule
\multirow{6}{*}{CV} & small & & - & $33.37$ \\
       & small & & GPC-50 & $\underline{16.4}$ \\
       & medium & & - & $21.3$ \\
       & medium & & GPC-50 &  $\mathbf{12.8}$ \\
       & large-v2 & & - & $16.03$ \\
       & large-v3 & & - & $13.98$ \\
\midrule
\multirow{6}{*}{LG} & small & & - & $30.35$ \\
       & small & & GPC-50 & $\underline{13.87}$ \\
       & medium & & - & $17.03$ \\
       & medium & & GPC-50 &  $\underline{11.84}$ \\
       & large-v2 & & - & $12.63$ \\
       & large-v3 & & - & $\mathbf{10.86}$ \\
\midrule
\multirow{6}{*}{HParl} & small & & - & $71.45$ \\
       & small & & GPC-50 & $\underline{32.35}$ \\
       & medium & & - & $39.27$ \\
       & medium & & GPC-50 &  $\underline{30.65}$ \\
       & large-v2 & & - & $27.09$ \\
       & large-v3 & & - & $\mathbf{16.99}$ \\
\bottomrule
\end{tabular}
\end{table}

\begin{figure*}[ht]
    \centering
    \includegraphics[width=\textwidth]{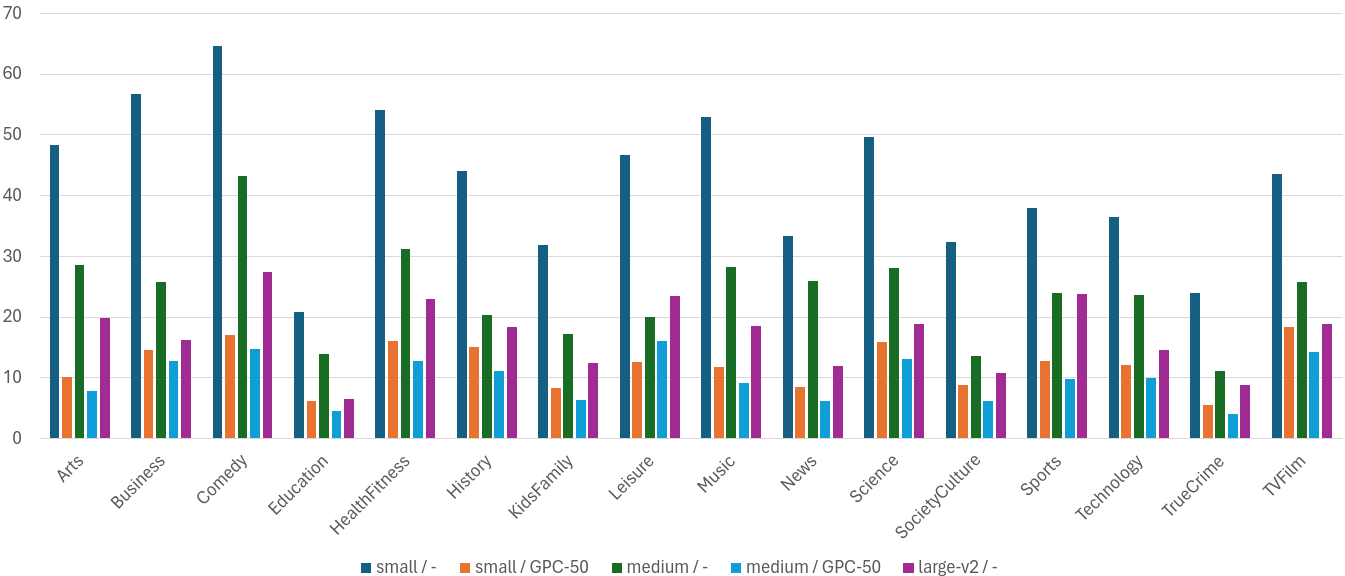}
    \caption{WER performance of whisper variant / fine-tuning corpus combinations over the different domains of the GPC-50 test set.}
    \label{fig:domain}
\end{figure*}

\begin{figure}
    \centering
    \includegraphics[width=\columnwidth]{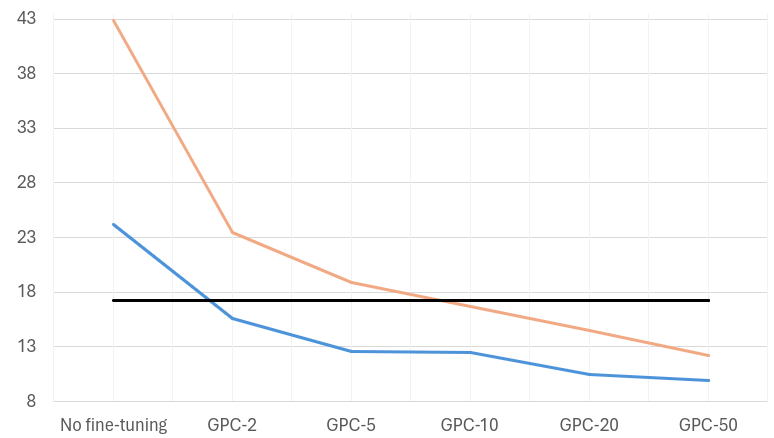}
    \caption{Performance of whisper-small (orange) and whisper-medium (blue) on the GPC-50 test set for varying amounts of data, when fine-tuning using the GPC-\{50,20,10,5,2\} training sets. The black line represents whisper-large-v2.}
    \label{fig:varying-data}
\end{figure}

\noindent\textbf{HParl}~\cite{10301554} is a corpus for Modern Greek, collected from the recorded proceedings of the Greek parliament. The corpus contains $50$ parliamentary sessions with $120$ hours of speech from $387$ speakers, 
and includes transcripts recorded in real-time by 
secretaries. The audio is segmented using an iterative alignment algorithm~\cite{manohar2017jhu}, implemented using 
Kaldi \cite{povey2011kaldi}. The average duration of the resulting segments is $4.6$ seconds. The acoustic conditions of HParl are challenging since large reverberation is present, due to the recording conditions in the parliament chamber. The HParl test set contains $11$ hours of speech.


\section{Results}

Our experiments revolve around three main research questions:
i) Does fine-tuning on pseudo-labeled corpora help with unseen datasets 
ii) What is the ASR performance across domains, and
iii) How does performance scale with increasing data size.

In Table~\ref{tab:podcast-50h}, we present the evaluation of whisper-{small, medium, large-v2}~\footnote{Whisper-large-v3 is not included in this evaluation, as it was utilized to create the GPC-50 transcriptions.} on the test set of the GPC-50 corpus. We assess the out-of-the-box Whisper models and the versions of Whisper small and medium fine-tuned on the GPC-50 training set. We observe a significant reduction in WER for the fine-tuned versions, with approximately 4x and 2x absolute improvements for small and medium, respectively. Both fine-tuned versions outperform whisper-large-v2.

Answering the first research question, we evaluate the fine-tuned models across four standard corpora, with results displayed in Table~\ref{tab:50h_dataset_model_overview}. We assess all Whisper variants from Table~\ref{tab:podcast-50h}, including the large-v3 version. Once again, fine-tuning results in significant, albeit more moderate, improvements compared to the pre-trained versions of Whisper-small and Whisper-medium. In the cases of Fleurs, CV, and LG, the fine-tuned models perform competitively with their larger counterparts, even surpassing them in the CV dataset.
For HP, we note that large-v3 exhibits a substantial performance advantage over the other variants, possibly due to its expanded training set, which may include speech with similar acoustic characteristics and/or employs reverberation-based augmentation.

In Fig.~\ref{fig:domain}, we evaluate the different 
Whisper versions across the $16$ domains in the GPC-50 test set. We observe that the fine-tuned versions consistently outperform whisper-large-v2 across all domains. Furthermore, we note variations in 
difficulty across domains. For instance, ``Comedy,'' which features laughter and conversational speech, appears to be more challenging. Conversely, ``Education'' and ``True Crime,'' characterized by slow, clear speech from a single speaker and better recording conditions, 
result in a WER in the range of $4$--$5$.

Finally, we fine-tune the small and medium variants of whisper on varying amounts of training speech and evaluate on the GPC-50 test set. The results can be seen in Fig.~\ref{fig:varying-data}. We verify that the performance scales with increasing the training corpus size, surpassing the large-v2 variant when we use the GPC-5 subset with $60$ hours of useable speech.


\section{Conclusions}

In this paper, we 
explored the impact of pseudo-transcriptions for training speech recognition models for low-resource languages.
For this, we 
collected an evolving corpus for Modern Greek from podcasts. The Greek Podcast Corpus has been transcribed with the WhisperX pipeline using the state-of-the-art large-v3 
model. Our evaluation aligns with current trends, showing that weak supervision can be a cost-effective strategy to create large corpora and train competitive models. This is especially relevant in the case of low-resource languages, where the amount of available data is the limiting factor.
In the future, we want to expand the GPC 
with more data, while also creating a gold split for reliable evaluation. Furthermore, we want to explore effective fine-tuning strategies and explore additional applications (e.g., speech-to-text translation and summarization).


\ifinterspeechfinal
\section{Acknowledgements}

Funded by the European Union. PREMIERE Project (No. 101061303). Views and opinions expressed are however those of the authors only and do not necessarily reflect those of the European Union.
\fi

\bibliographystyle{IEEEtran}
\bibliography{mybib}

\end{document}